\documentclass[sigconf]{acmart}
\usepackage{bm}
\usepackage{graphicx}
\graphicspath{ {images/} }
\usepackage{color}

\AtBeginDocument{%
  \providecommand\BibTeX{{%
    \normalfont B\kern-0.5em{\scshape i\kern-0.25em b}\kern-0.8em\TeX}}}

\setcopyright{iw3c2w3}
\copyrightyear{2021}
\acmYear{2021}
\acmDOI{}

\acmConference[DeMaL@WWW '21]{DeMaL@WWW '21: WWW Workshop on Data-Efficient Machine Learning for Web Applications}{April 19--23, 2021}{Ljubljana, Slovenia}
\acmBooktitle{}
\acmPrice{}
\acmISBN{}

\begin{document}

\title{Multimodal and Contrastive Learning for Click Fraud Detection}


\author{Weibin Li*, Qiwei Zhong*, Qingyang Zhao, Hongchun Zhang, Xiaonan Meng}
%\thanks{*Corresponding author.}
\authornote{Corresponding author.}
\affiliation{%
  \institution{Alibaba Group}
  \city{Hangzhou}
  \country{China}}
\email{{dece.lwb, yunwei.zqw, qingyang.zqy, hongchun.zhc, xiaonan.mengxn}@alibaba-inc.com}

\renewcommand{\shortauthors}{Weibin Li and Qiwei Zhong, et al.}

\begin{abstract}
Advertising click fraud detection plays one of the vital roles in current E-commerce websites as advertising is an essential component of its business model.
It aims at, given a set of corresponding features, e.g., demographic information of users and statistical features of clicks, predicting whether a click is fraudulent or not in the community.
Recent efforts attempted to incorporate attributed behavior sequence and heterogeneous network for extracting complex features of users and achieved significant effects on click fraud detection.
In this paper, we propose a \underline{M}ultimodal and \underline{C}ontrastive learning network for \underline{C}lick \underline{F}raud detection (MCCF).
Specifically, motivated by the observations on differences of demographic information, behavior sequences and media relationship between fraudsters and genuine users on E-commerce platform, MCCF jointly utilizes wide and deep features, behavior sequence and heterogeneous network to distill click representations.
Moreover, these three modules are integrated by contrastive learning and collaboratively contribute to the final predictions.
With the real-world dataset containing 3.29 million clicks on Alibaba platform, we investigate the effectiveness of MCCF.
The experimental results show that the proposed approach is able to improve AUC by 7.2\% and F1-score by 15.6\%, compared with the state-of-the-art methods.
\end{abstract}

\iffalse

\fi



\maketitle
\section{Introduction}
Online click advertising, widely known as cost-per-click or pay-per-click, is an internet advertising in which an advertiser pays a publisher (typically a search engine, website owner, or a network of websites) when one ad is clicked~\footnote{https://en.wikipedia.org/wiki/Pay-per-click}.
Different from traditional advertising, advertisers can track consumers' online behaviors for accurate measurements of advertising profitability~\cite{wilbur2009click}.
Click fraud detection plays a critical role due to the growing volume of this online advertising.
Google implicitly acknowledged the problem when it paid \$90 million to settle a click fraud lawsuit~\cite{tuzhilin2006lane}.
Moreover, the World Federation of Advertisers says ad fraud will cost advertisers \$50 billion a year by 2025~\footnote{https://www.businessinsider.com/wfa-report-ad-fraud-will-cost-advertisers-50-billion-by-2025-2016-6}.
 \begin{figure*}
    \centering
    \includegraphics[width=17cm]{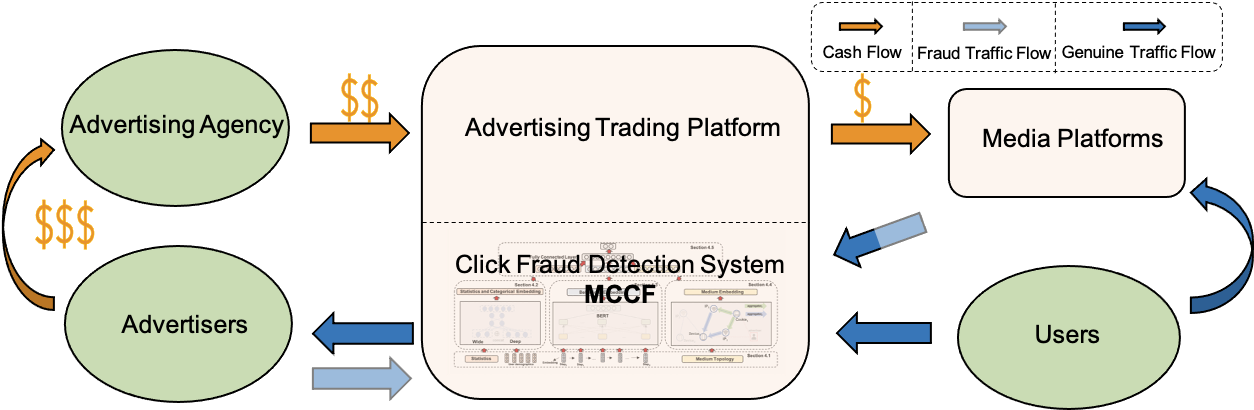}
    \caption{A typical flow of click advertising business.}
    \label{fig:ad}
\end{figure*}

Practically, click advertising is sold on per click basis.
Figure~\ref{fig:ad} shows the four roles in the typical advertising business scenario. Their functions, interest appeals and click fraud motivations are summarized as follows:
\begin{itemize}
\item Advertisers: reaching users with advertisements of their products, and further converting users to consumers of their services or products. Advertisers may click rivals' ads with the purpose of driving up their costs or exhausting their ad budget. When a rival's budget is exhausted, it will exit the ad auction.
\item Advertising Agency: more professional advertising promotion trader, helping advertisers manage their accounts and providing professional marketing services. They have no incentive for click fraud.
\item Advertising Trading Platform: advertising platform that connects internet media and advertisers. It not only provides advertisers with advertising marketing tools and advertising services, but also realizes the commercial value of advertising with the help of internet media traffic. For example, search engine companies, e-commerce companies, and social companies with a large number of users and traffic. They have no incentive for click fraud as well.
\item Media Platforms: providers of internet information and services. When users browse their information or use their services, they complete the dissemination of advertising information. The media is generally also called an alliance, such as blogs and address navigation websites. Some of these third parties might click the ads maliciously to inflate advertisers' revenues.
\item Users: person who browses information or uses services on the internet is a potential customer of an advertiser. They also have no incentive for click fraud.
\end{itemize}

Although the existing researches have achieved significant effects in the detection of common frauds such as machine click fraud or click fraud with distinct statistical features~\cite{antoniou2011exposing,badhe2017click,faou2016follow,haddadi2010fighting,kshetri2010economics,mouawi2018towards,thejas3hybrid,thejas2019deep,xu2014click}, the detection of high-level fraud still needs to be resolved. The particular challenges of this issue are summarized as follows:
\begin{itemize}
\item Simulate genuine click behavior: fraudsters simulate genuine click, manifesting as more complex abnormality of statistical features.
\item Fraudsters frequently switch IP and clear cookies to make their statistical features look like genuine. However, their behavior sequence might be abnormal, such as only visiting search and advertising pages.
\item Group fraud involving heterogeneous information: a group of multiple people attack a specific advertiser together.
\item Highly imbalanced distribution: the ratio of fraudulent clicks to genuine clicks is less than 1:8 for instance.
\end{itemize}

Therefore, building a more effective fraud detection system is pivotal for online advertising businesses. Specifically, based on the challenges above as well as the analysis and observations below on the real-world dataset, we propose a novel \underline{M}ultimodal and \underline{C}ontrastive learning network for \underline{C}lick \underline{F}raud detection (\textbf{MCCF}). %~\footnote{Our code and data would be disclosed if the paper is luckily accepted.}
Firstly, multimodal information including statistic and categorical features, behavior sequences and media relationships modeled by Wide and Deep~\cite{cheng2016wide}, BERT~\cite{devlin2018bert,vaswani2017attention} and GNN~\cite{hamilton2017inductive,liang2021credit,liu2019geniepath,wu2020comprehensive,zhang2020deep,zhong202020mahin,zhou2018graph} are involved to perform comprehensive click representations simultaneously.
Secondly, we integrate these representations via multiple layer perceptron and output the prediction. Finally, contrastive learning~\cite{chen2020simple} is utilized to solve the imbalance problem in this domain.
%"Click fraud" is the practice of deceptively clicking with the purpose of  either increasing third-party website undue monetary returns or exhausting an advertiser's budget \cite{jansen2007click}. The World Federation of Advertisers says ad fraud will cost advertisers \$50 billion a year by 2025. It's second only to the drugs trade as a source of income for organized crime. Most academics and consultants estimate that 10 to 20 percent of ad clicks are fake \cite{kshetri2010economics}. Google implicitly acknowledged the problem when it paid \$90 million to settle a click-fraud lawsuit  in July 2006 \cite{tuzhilin2006lane}. Survey found that 42\% of advertisers are victims of click fraud.Kenneth \cite{wilbur2009click} find that if x\% of clicks are fraudulent, advertisers will lower their bids by x\%, leaving ad platform revenues unchanged.
%Table \ref{table:clickField} lists the fields in the click database.Figure 2(a) below is a list of search results on the mobile terminal of an e-commerce website, with ads at the top of the search results. Figure 2(b) shows the basic attributes of a click (such as IP, Query, Company), the clicker’s media information (such as IP, CookieID, DeviceID), and the clicker’s sequence of actions within the website.
\begin{table}[]
    \centering
\caption{Some typical fields in click fraud detection system.}
    \resizebox{\linewidth}{!}{
\begin{tabular}{ c | c }
 \hline
\textbf{Field}      & \textbf{Description}                             \\
 \hline
 AbsPos     & Absolute position of an ad  on website\\
 AdvertiserID & Unique identifier of advertiser               \\
 CdTime     & Interval between display time and click time  \\
ClickID    & Unique identifier of a particular click   \\
ClickTime  & Timestamp of a given click                \\
CookieID     & Unique identifier of users             \\
CookieTime & Timestamp that cookie was generated    \\
DeviceID     & Unique identifier of mobile users  \\
IP         & Public IP address of a click            \\
KeywordID     & Unique identifier of ad word  \\
PageType   & Homepage, Detail, ... \\

\hline 
 
\end{tabular}
}
 
 \label{table:clickField}
 \vspace{-0.2cm}
\end{table}

\vspace{0.5em}\textit{Observation 1: The statistical feature of clicks are clearly distinct between genuine users and fraudsters.}

\begin{figure*}
    \centering
    \includegraphics[width=16.8cm]{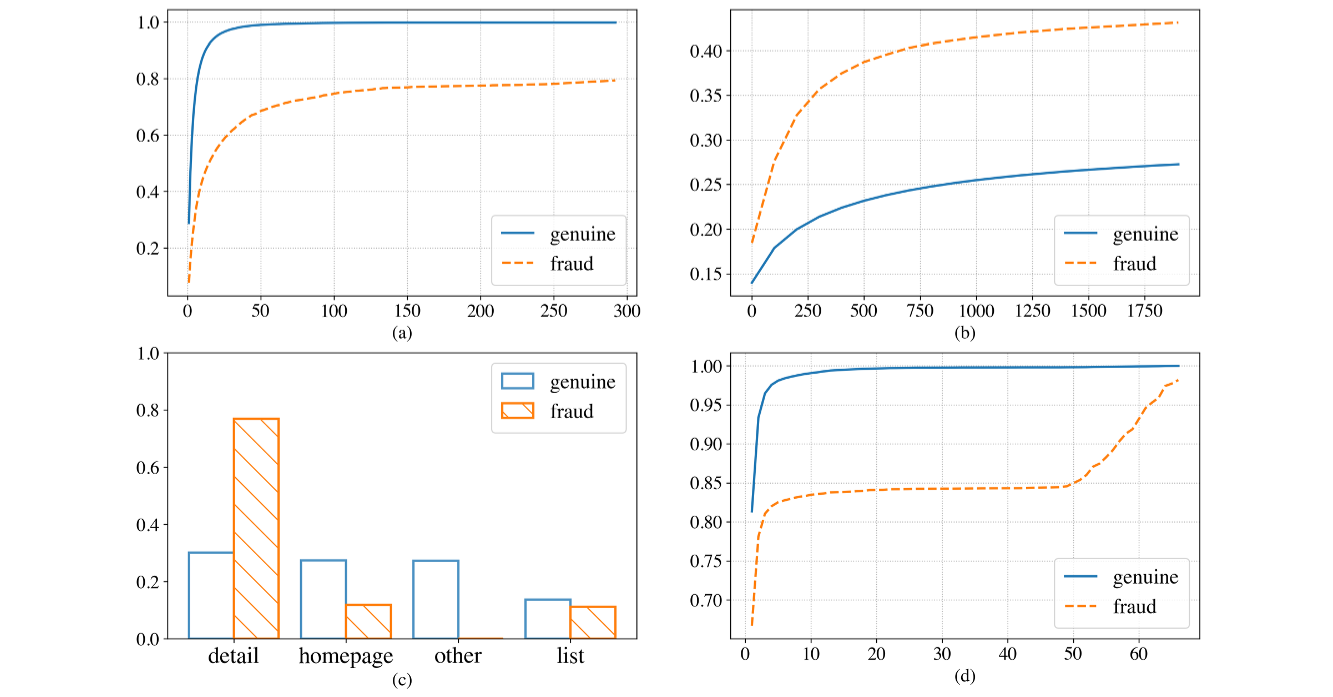}
    \caption{Statistical feature of fraudsters and genuine users: (a) cumulative distributions of the average number of clicks per IP; (b) cumulative distributions of the average time interval between the click time and the time that CookieID was generated; (c) page categorical properties of behaviors; (d) cumulative distributions of the average number of media.}
    \label{fig:analysis}
     %\vspace{-0.1cm}
\end{figure*}

Figure~\ref{fig:analysis} (a) and (b) illustrate the cumulative distributions of the average number of clicks per IP per day, and the average time interval between the click time and the time that CookieID was generated for genuine and fraud clicks on Alibaba.com, respectively. We found that the number of clicks per IP of most fraudsters in a single day is much more than that of genuine users. For example, 54.69\% of fraud clicks have at least 10 times on their number of clicks per IP, while only 11.53\% for genuine clicks. Meanwhile, we observed that time interval between the click time and the time that CookieID was generated for fraudsters are much shorter, e.g., 40.81\% v.s. 24.78\% of the intervals are $\leq 900$ seconds for fraudsters and genuine users, respectively. We can easily conjecture the reason is that fraudsters try to fraudulently click as many as possible for a better ROI~\footnote{https://en.wikipedia.org/wiki/Return\_on\_investment}.

\vspace{0.5em}\textit{Observation 2: The difference of behavior pages between genuine and fraud clickers are significant.}

As shown in Figure~\ref{fig:analysis} (c), we demonstrate the ratio of top page types between fraud and genuine clicks. For example, over 99\% of fraudsters are concentrated on homepage, detail, and list pages, while the proportion of genuine clicks on each page is relatively even.

\vspace{0.5em}\textit{Observation 3: Both number of associated media are distinguished between genuine users and fraudsters.}

Figure~\ref{fig:analysis} (d) illustrates the cumulative distributions of the average number of media (such as IP, CookieID, DeviceID) of clicks from fraudsters and genuine users. We clearly observe that the number of associated media of most fraudsters in a single day is much more than that of genuine users, which results in flatter trends on the corresponding cumulative distribution curve. For example, 21.86\% of fraudsters have at least 3 associated medias, while only 6.31\% for genuine users.

%Building an effective fraud detection system is thus pivotal for online advertising businesses. Based on data analysis, we proposed a system called multimodal contrastive learning for click fraud detection(MCCF). 
The main contributions of this work are summarized as follows: 
\begin{itemize}
\item To the best of our knowledge, we are the first attempt to incorporate multimodal information and contrastive learning for click fraud detection.
\item We propose a novel multimodal and contrastive learning network to solve this problem. Specifically, multimodal information including statistic and categorical features, behavior sequences and media relationships are involved to perform comprehensive click representations, and multiple layer perceptron is utilized to integrate them. Furthermore, to solve the imbalance problem, contrastive learning is elaborately exploited during training.
\item Experiments on real-world dataset demonstrate the effectiveness of the proposed approach. It achieves competitive performance and outperforms state-of-the-art methods.
\end{itemize}

\section{Related work}
%Kshetri~\cite{kshetri2010economics} classified click fraud detection methods into three categories: anomaly-based, rule-based and classifier-based. The anomaly-based approach considers invalid clicks to be those that deviate significantly from normal predicted behaviors. The rule-based approach uses heuristics to classify valid and invalid clicks on the basis of specific conditions. For instance, if two successive clicks occur, the second click might be an invalid one. Finally, the classifier-based approach is purely operational and employs data mining classifier labels to detect invalid clicks.
In this section, we review related studies from three aspects, namely avoid click fraud in advance, anomaly-based and rule-based methods, and classifier-based methods. These related researches are categorized as follows:
\subsection{Avoid click fraud in advance}
Haddadi~\cite{haddadi2010fighting} presents bluff ads, a strategy to increase the effort of click fraudsters. CAPTCHA is used to ensure that the click is legitimate ~\cite{costa2012proposal,thejas2019deep}. Faou~\cite{faou2016follow} follows the traffic to stop click fraud by disrupting the value chain. These methods increase the cost of click fraud, and meanwhile it may hurt the user experience to a certain extent.

\subsection{Anomaly-based and Rule-based methods}
Kshetri~\cite{kshetri2010economics} classifies click fraud detection methods into three categories: anomaly-based, rule-based and classifier-based.
Antoniou and Zhang~\cite{antoniou2011exposing,zhang2008detecting} analyze the number of visits in a certain time interval to detect duplicate clicks. Badhe~\cite{badhe2017click} uses programmatic scripts to detect machine click fraud. Kitts~\cite{kitts2008identifying} devises algorithm to detect robot click fraud. Due to the strong interpretability of the rules, Kitts~\cite{kitts2015click} uses rules to filter click fraud early. But as fraud escalates, the rules become difficult to maintain and the detection ability deteriorates.

\subsection{Classifier-based methods}
Xu~\cite{xu2014click} constructs a pruned decision tree to classify traffic as valid, casual or fraudulent and introduces additional tests to check whether visiting clients are click-bots. Mouawi and Oentaryo~\cite{mouawi2018towards,oentaryo2014detecting} present an important application of machine learning and data mining methods to tackle click fraud detection problems, such as single algorithms (e.g., LR, SVM, kNN, ANN) and ensemble learning algorithms (e.g., Random Forest).
Kitts~\cite{kitts2015click} discusses how to design a data mining system to detect large scale click fraud attacks.
Berrar, Minastireanu and Oentaryo~\cite{berrar2012random,minastireanu2019light,oentaryo2014detecting} prove that LightGBM and Random Forest have achieved good results.
Thejas~\cite{thejas3hybrid} combines Cascaded Forest and XGBoost to detect click fraud.
Perera~\cite{perera2013novel} utilizes an ensemble method to detect click fraud, which gained higher performance than single classifiers.
Thejas~\cite{thejas2019multi,thejas2019deep} proposes a hybrid deep learning model consisting of an Auto Encoder, a Neural Network and a Semi-supervised Generative Adversarial Network (GAN) to predict click fraud in imbalanced dataset.
Although the above models can recall some fraud, they cannot effectively detect advanced fraud and group fraud that simulate genuine user behaviors.
 \begin{figure*}
    \centering
    \includegraphics[width=\linewidth]{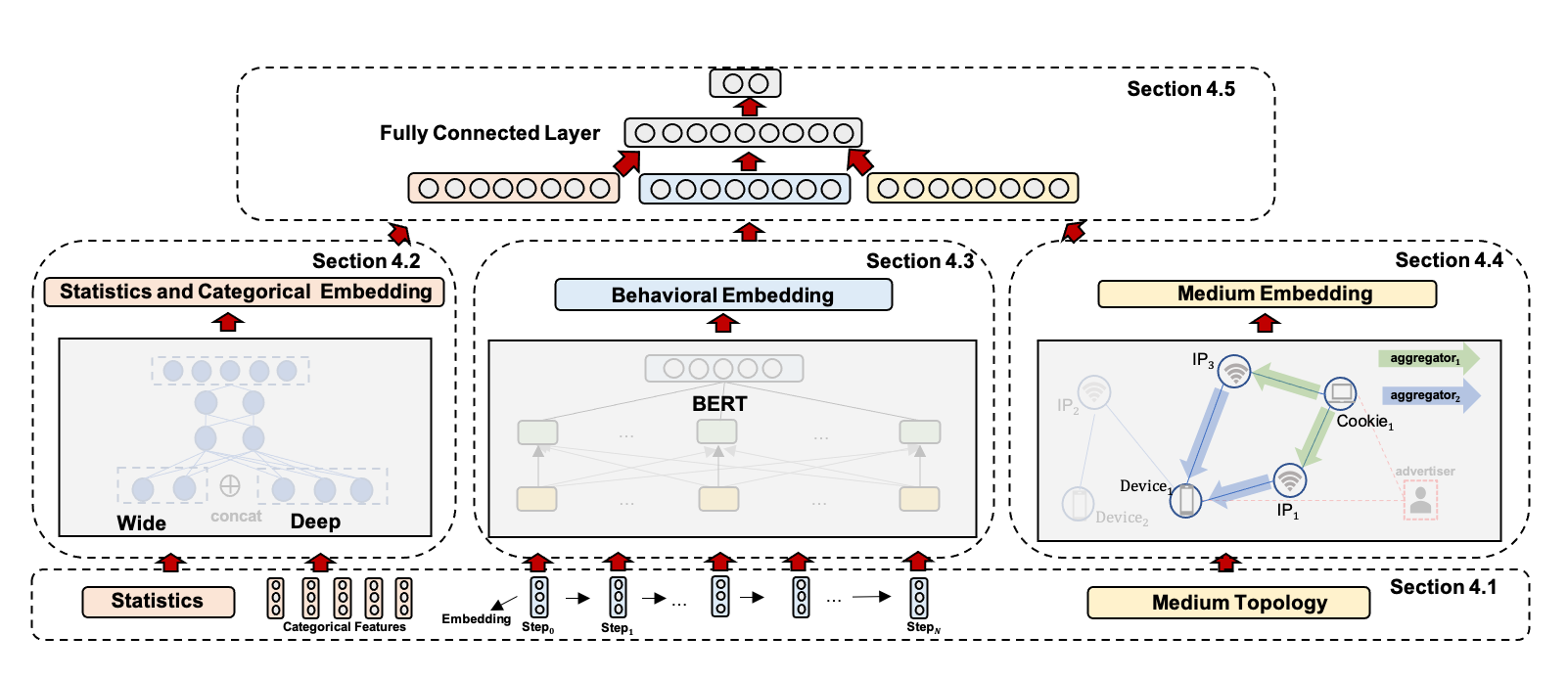}
    \caption{An illustration of the proposed MCCF model.}
    \label{fig:model structure}
\end{figure*}

\section{PROBLEM STATEMENT}
In this section, we present the problem formulation for click fraud.
A click $\bm{x}$ in our problem consists of three kinds of information, namely Wide and Deep feature (denoted as $\bm{x}^{(w)}$, $\bm{x}^{(d)}$), behavior sequence (denoted as $\bm{x}^{(b)}$) and graph feature of user (denoted as $\bm{x}^{(v)}$). Given a set of the corresponding features, the goal of this task aims at predicting whether the click is fraudulent or not. Prior to that, we introduce several definitions which are helpful for problem statement.

\vspace{0.5em}\noindent \textbf{Definition 1. Wide and Deep feature:}
wide features are continuous features in each click, including original values (e.g., CdTime), combined features (e.g., AbsPos and CdTime), and demographic features (e.g., the number of cookies in the last day of IP). Deep features are categorical features in each click, such as AdvertiserID, KeywordID. $\bm{x}^{(w)} = [x^{(w)}_1, x^{(w)}_2, ..., x^{(w)}_l]$ is a vector of $l$-dimensional wide feature, and $\bm{x}^{(d)} = [x^{(d)}_1, x^{(d)}_2, ..., x^{(d)}_r]$ is a vector of $r$-dimensional deep feature.

\vspace{0.5em}\noindent \textbf{Definition 2. Behavior sequence:}
the sequence of pages visited by a user before the ad is clicked, such as ``Homepage -> List -> Detail -> $\cdots$'', as shown in Figure~\ref{fig:model structure}. $\bm{x}^{(b)} = [x^{(b)}_1, x^{(b)}_2, ..., x^{(b)}_t]$ is a vector of $t$-dimensional behavior sequence. Specifically, the value of $t$ in our model is 300.

\vspace{0.5em}\noindent \textbf{Definition 3. Multi-media heterogeneous network:}
given a graph $G = (V, E)$, the feature of the node $\bm{x}^{(v)} = [x^{(v)}_1, x^{(v)}_2, ..., x^{(v)}_s]$ is a vector of $s$-dimensional feature integrated from itself and neighbors. The node types in our heterogeneous network are IP, CookieID, and DeviceID. For example, If a CookieID uses an IP to visit the website, the two nodes are neighbors, and an edge will be connected between them (as shown in Figure~\ref{fig:model structure}).
For attributes of heterogeneous network, we collect 542 attributes for each medium (node), such as demographic information and click frequency. For each relation (link), we construct 90 attributes such as link type (e.g., click, login, and pay), first/last related time, and interaction frequency.

\section{THE MCCF MODEL}
In this section, we present the proposed MCCF model, as shown in Figure~\ref{fig:model structure}. We firstly introduce the distilling of feature representations and then illustrate model training via contrastive learning.
%wide and deep features that characterize machine fraud and common human fraud, and then introduce the extraction of behavior sequence fraud feature, we introduce the extraction gang fraud feature, and finally describe our model training using Contrastive Learning as the loss function.
\subsection{Input layer}
Every element in the sequences $\bm{x}^{(d)}$ and $\bm{x}^{(b)}$ for each click needs to be transferred into embedding. After looking up from two embedding matrices $\bm{W}^{(d)}$, $\bm{W}^{(b)}$ respectively, $\bm{x}^{(d)}$ and $\bm{x}^{(b)}$ are converted to $\bm{e}^{(d)} = [\bm{e}^{(d)}_1, \bm{e}^{(d)}_2, ..., \bm{e}^{(d)}_r]$, $\bm{e}^{(b)} = [\bm{e}^{(b)}_1, \bm{e}^{(b)}_2, ..., \bm{e}^{(b)}_t]$, of which each element is an embedding vector, as shown in Figure~\ref{fig:model structure}.
\begin{equation}\bm{e}^{(d)} =\bm{\mathrm{LOOKUP}}\left(\bm{W}^{(d)};\bm{x}^{(d)}\right)\label{equ:deep input embedding}\end{equation}
\begin{equation}\bm{e}^{(b)} =\bm{\mathrm{LOOKUP}}\left(\bm{W}^{(b)};\bm{x}^{(b)}\right)\end{equation}
where $\bm{\mathrm{LOOKUP}}\left(\bm{W};\bm{x}\right)$ is an operator to get vectors from $\bm{W}$ using each element of $\bm{x}$ as subscript. 
%We denote all parameters ($\bm{W}^{(d)}$, $\bm{W}^{(b)}$) as $\Theta^1$. 
The embedding vectors are initialized randomly and then the values are trained with the model parameters to minimize the final loss function during training.

\subsection{Wide and Deep Network}
The wide and deep components are a multilayer neural network, as shown in Figure~\ref{fig:model structure}. The original inputs of deep component are categorical features (e.g., AdvertiserID, KeywordID). Each of these sparse, high-dimensional categorical features $\bm{x}^{(d)}$ are converted into a low-dimensional and dense embedding vector $\bm{e}^{(d)}$ via equation~(\ref{equ:deep input embedding}). These low-dimensional dense embedding vectors $\bm{e}^{(d)}$ and wide feature $\bm{x}^{(w)}$ are concatenated and then fed into the hidden layers of a neural network in the forward pass. Specifically, the wide and deep components perform the following computation.
\begin{equation}\bm{e}^{(wd)} = \bm{\mathrm{CONCAT}}\left(\bm{e}^{(d)},\bm{x}^{(w)}\right)\end{equation}
\begin{equation}\bm{v}^{(wd)} = \bm{\mathrm{ReLU}}\left(\bm{W}_{wd}^{(L)}…\bm{\mathrm{ReLU}}\left(\bm{W}_{wd}^{(1)}\bm{e}^{(wd)}+\bm{b}_{wd}^{(1)}\right) + \bm{b}_{wd}^{(L)}\right)\end{equation}
where $L$ is the layer number, $\bm{W}_{wd}^{(l)}$ and $\bm{b}_{wd}^{(l)}$ are the model weights and bias at $l^\mathrm{th}$ layer. $\bm{v}^{(wd)}$ is the wide and deep component embedding vector.

\subsection{Behavior Sequence Network}
%Recently, LSTM~\cite{hochreiter1997long} and TextCNN~\cite{zhang2015sensitivity} have achieved good results in text classification, but Bert~\cite{devlin2018bert} has achieved SOTA in text classification tasks.
We utilize BERT~\cite{devlin2018bert} which gets SOTA results on many tasks to model behavior sequence, as shown in Figure~\ref{fig:model structure}.
%The BERT model is a language model based on the bidirectional Transformer. As shown in Figure \ref{fig:model structure}. 
%Transformer encoder includes multi-layer Multi-Head Attention and feed forward neural network. The self-attention mechanism enables the current node to pay attention to the current word and obtain the semantics of the context. Multi-Head Attention expands the model's ability to focus on different locations. 
For the input embedding vector $\bm{e}^{(b)}$, BERT converts it into representation vector $\bm{v}^{(b)}$, paying more attention to the page type that can distinguish between fraud and genuine click.
%different page types. During training, Bert focuses his attention . We denote all parameters on Bert as $\Theta3$. 
\begin{equation}\bm{v}^{(b)} = \bm{\mathrm{BERT}}\left(\bm{e}^{(b)}\right)\end{equation}

\subsection{Multi-media Heterogeneous Network}
The core idea of a multi-media heterogeneous network is to aggregate the neighbors' feature information. As shown in Figure~\ref{fig:model structure}, the fraudster may exchange multiple cookies and devices of different media for click fraud. After the aggregating of statistical features of neighbor nodes via media heterogeneous network, fraudster's feature distribution might be quite abnormal.
\begin{equation}
\bm{h}_{\mathcal{N}(v)}^{k} = \bm{\mathrm{AGGREGATE }}_{k}\left(\left\{\bm{h}_{u}^{k-1}, \forall u \in \mathcal{N}(v)\right\}\right)
\end{equation}
\begin{equation}
\bm{h}_{v}^{k} = \sigma\left(\bm{\mathrm{W}}_v^{k} \cdot \bm{\mathrm{CONCAT}}\left(\bm{h}_{v}^{k-1}, \bm{h}_{\mathcal{N}(v)}^{k}\right)\right)
\label{equ:graph embedding}
\end{equation}
where $\bm{h}_{v}^{k}$ denotes a node’s representation at this step. $\mathcal{N}(v)$ are all neighbor nodes of node $v$. Note that this aggregation step depends on the representations generated at the previous iteration, and representation $\bm{h}_{v}^{0}$ = $\bm{x}^{(v)}$ is defined as the input node features. We use mean aggregation function here. Our method firstly aggregates the feature vector of the previous step of the neighbor node,  then concatenates the node’s current representation $\bm{h}_{v}^{k-1}$ as shown in equation~(\ref{equ:graph embedding}),
where $\sigma$ is nonlinear activation function and $k$ is the depth of the search. For notation convenience, we denote the final output representation at depth $k$ as $\bm{v}^{(v)}$ = $\bm{h}_{v}^{k}$.

\subsection{Integration and Training}
Finally, the outputs of the three modules are concatenated, followed by two fully connected layers and an output layer based on $\mathsf{softmax}$ function, as shown in Figure \ref{fig:model structure}. The concatenation is represented as $\bm{v}^{(i)}=\left[\bm{v}^{(wd)}; \bm{v}^{(b)}; \bm{v}^{(v)}\right]$ and
the next layers are denoted as
%\begin{equation}
%\bm{z}_{1}^{(i)}=\bm{\mathrm{ReLU}}\left(\bm{W}_{1}^{(i)} \bm{v}^{(i)}+\bm{b}_{1}^{(i)}\right)
%\end{equation}
\begin{equation}
%\bm{z}_{2}^{(i)}=\bm{W}_{2}^{(i)} \bm{z}_{1}^{(i)}+\bm{b}_{2}^{(i)}
\bm{z}_{2}^{(i)}=\bm{W}_{2}^{(i)} \left(\bm{\mathrm{ReLU}}\left(\bm{W}_{1}^{(i)} \bm{v}^{(i)}+\bm{b}_{1}^{(i)}\right)\right)+\bm{b}_{2}^{(i)}
\end{equation}
\begin{equation}
\hat{\bm{y}}=\bm{\mathrm{softmax}}\left(\bm{z}_{2}^{(i)}\right)
\end{equation}
where $\bm{W}_{k}^{(i)}, \bm{b}_{k}^{(i)} (k=1,2)$ are weight matrix and bias vector of each layer, and $\bm{\mathrm{ReLU}}(\cdot)$ is the element-wise rectified linear unit function. Specifically, $\hat{\bm{y}}$ is defined as the predicted probability vector of a click.

Furthermore, as mentioned previously, there is highly imbalance problem in click fraud detection task generally. To solve it, contrastive learning is elaborately exploited during training.
Hadsell, Chopra, and LeCun~\cite{hadsell2006dimensionality} propose a loss function coined max margin contrastive loss that operates on pairs of samples instead of individual samples. Intuitively, this loss function learns an embedding to place samples with the same labels close to each other, while distancing the samples with different labels. Weinberger and Sohn~\cite{sohn2016improved,weinberger2009distance} present a multi-class N-pair loss which is an upgrade of max margin contrastive loss allowing joint comparison among more than one negative samples. Chen and Khosla~\cite{chen2020simple,khosla2020supervised} propose the normalized temperature-scaled cross entropy loss (NT-Xent). It is a modification of multi-class N-pair loss with addition of the temperature parameter ($\tau$).

In this paper, we train our model with SOTA NT-Xent loss with regularization.
Specifically, let multiple layer perceptron ($\bm{\mathrm{MLP}}$) be an encoder network mapping $\bm{z}_{2}^{(i)}$ to the latent space $\bm{z}$ firstly.
\begin{equation}\bm{z} = \bm{\mathrm{MLP}}\left(\bm{z}_{2}^{(i)}\right)\end{equation}

Let $\mathsf{sim}(\bm{a}, \bm{b})$ denote the dot product between $\ell_{2}$ normalized $\bm{a}$ and $\bm{b}$ (i.e. cosine similarity) in equation~(\ref{equ:cosine similarity}).
When applied on a pair of positive samples $\bm{z}^{(i)}$ and $\bm{z}^{(j)}$ and other $2(N-1)$ negative examples, the loss function $\ell(i,j)$ for a positive pair of examples $(i,j)$ is defined in equation~(\ref{equ:pair loss function}).
\begin{equation}
\mathsf{sim}(\bm{a}, \bm{b})=\bm{a}^{\top} \bm{b} /\|\bm{a}\|\|\bm{b}\|
\label{equ:cosine similarity}
\end{equation}
\begin{equation}
\ell(i,j)=-\log \frac{\exp \left(\mathsf{sim}\left(\bm{z}^{(i)}, \bm{z}^{(j)}\right) / \tau\right)}{\sum_{k=1}^{2N} \mathbb{1}_{[k \neq i]} \exp \left(\mathsf{sim}\left(\bm{z}^{(i)}, \bm{z}^{(k)}\right) / \tau\right)}
\label{equ:pair loss function}
\end{equation}
where $\mathbb{1}_{[k \neq i]} \in\{0,1\}$ is an indicator function evaluating to 1 if $k \neq i$, and $\tau$ denotes a temperature parameter.
The final loss is computed across all positive pairs in a mini-batch.
%This loss has been used in previous work \cite{khosla2020supervised} , and shown in Equation~\ref{equ:final loss}. 
\begin{equation}
\mathcal{L}=\frac{1}{2 M} \sum_{k=1}^{M}\left[\ell(2 k-1,2 k)+\ell(2 k, 2 k-1)\right] +\frac{\lambda}{2}\|\bm{\mathrm{\theta}}\|_{2}^{2}
\label{equ:final loss}
\end{equation}
where $\lambda$ is the regularization parameter and $\bm{\mathrm{\theta}}$ is the set of parameters of the proposed model.

\subsection{Discussions}
It is worth noting that not all kinds of sequences or networks mentioned above are compulsory in our MCCF model. For situations where only part features are available, it works as well. It can accomplish prediction by switching off the corresponding parts in the integration stage. For example, we can use only wide and deep sequences for early detection of click fraud. The corresponding experimental results will be demonstrated in the following sections.

\section{EXPERIMENTS}
In this section, we investigate the effectiveness of the proposed model. We conduct extensive experiments on a large-scale real-world dataset. Firstly, we verify the performance on detecting frauds from the dataset. Secondly, we perform ablation test and visualization to demonstrate the effectiveness of every component in our model.

\subsection{Dataset}
We collect a real-world dataset from an online click advertising service on Alibaba.com under the premise of complying with security and privacy policies.
It contains 2.54 million clicks for training and 0.75 million clicks for testing, chronologically.
User's rich behavioral information such as clicking logs, media relationship logs are collected according to their chronological orders. Based on the dataset, we construct a multimodal attributed information network.
As mentioned previously, three modals are adopted, namely wide and deep features, behavior sequence and multi-media heterogeneous network, as shown in Figure~\ref{fig:model structure}.
It is worth noting that the label (fraud or genuine) of training and testing set are acquired via partially forecasting beforehand by the high-precision models deployed online, and manually evaluating and double checking offline afterwards.
%The positive rate is around 10.89\% in the dataset.
The data statistical information is exhibited in Table~\ref{table:statistical information}.
\begin{table}
\caption{The statistical information of dataset.}
\resizebox{\linewidth}{!}{
\begin{tabular}{c|c|c|c|c}
\hline \textbf{Dataset} & \textbf{\#Positive} & \textbf{\#Negative} & \textbf{\#Total} & \textbf{\#Positive Rate} \\
\hline Training & 276,956 & 2,265,022 & 2,541,978 & 10.89\% \\
Testing & 75,999 & 670,721 & 746,720 & 10.17\% \\
\hline
\end{tabular}
}
 
 \label{table:statistical information}
\end{table}
\subsection{Compared Methods}
We compare with several state-of-the-art representative methods including tree-based, graph-based and sequence-based to verify the effectiveness of our proposed method. Among them, tree-based baselines use statistical features, graph-based method uses medium topology information and statistical features, and the rests use behavior sequence information.

\noindent(a) \textbf{Tree-based Methods}
\begin{itemize}
\item \textbf{Random Forest}~\cite{berrar2012random,oentaryo2014detecting}: a scalable tree-based model for feature learning and classification task, and widely used in various areas.
\item \textbf{LightGBM}~\cite{ke2017lightgbm,minastireanu2019light}: an efficient parallel training Gradient Boosting Decision Tree-type method. Random Forest and LightGBM use statistical features, such as the number of cookies in the last day of IP and CdTime.
\end{itemize}
\noindent(b) \textbf{Graph-based Method}
\begin{itemize}
\item \textbf{GraphSAGE}~\cite{hamilton2017inductive}: a general and inductive framework that efficiently generates node embeddings by sampling and aggregating features from a node's local neighbors.
\end{itemize}
\noindent(c) \textbf{Sequence-based Methods}
\begin{itemize}
\item \textbf{BiLSTM}~\cite{hochreiter1997long}: it mines the contextual information of the behavior sequence, and uses the attention mechanism to extract important information, so as to realize the classification of the sequence.
\item \textbf{TextCNN}~\cite{zhang2015sensitivity}: an algorithm that uses convolutional neural networks to classify text sequence. Different convolutions are used to extract the features of the context at different local locations to obtain semantic information at different levels of abstraction.
\item \textbf{BERT}~\cite{devlin2018bert}: a pretrained model uses the now ubiquitous transformer architecture.
\end{itemize}
\noindent(d) \textbf{Our Method and Variants}
\begin{itemize}
\item \textbf{MCCF}: our proposed method. We also derive four variants of MCCF to comprehensively compare and analyze the performances of its each component. They are:
\item \textbf{MCCF}$_{\backslash WD}$: removing wide and deep features.
\item \textbf{MCCF}$_{\backslash B}$: removing behavior sequence.
\item \textbf{MCCF}$_{\backslash V}$: removing multi-media heterogeneous network.
\item \textbf{MCCF}$_{CE}$: changing the loss function from NT-Xent to cross entropy~\cite{rubinstein1999cross}.
\end{itemize}

% Several representative click fraud detection methods are compared in our evaluations. Among them, two baselines use statistical features, one method use medium topology information and statistical features, and the rests use behavioral sequence information.
\subsection{Implementation Details}
For the network structure, the size of wide feature is set to 40, the embedding vector for the input layer of deep feature is set to 128, and the input embedding of behavior sequence is set to 128. For the multi-media heterogeneous network, the aggregating function is mean, the depth of search is set to 2, and the size of node feature vector is set to 500. For training parameters, $\lambda$ is set to 0.01, the learning rate is set to 0.001, and the batch size is set to 64. We randomly initialize the model parameters with an xavier initializer~\cite{glorot2010understanding} and choose Adam~\cite{kingma2014adam} as the optimizer. Five-run-average values are reported.

Our experiment uses Precision, Recall, micro F1-Score and AUC to compare the effects of all methods. The higher these metrics indicate the higher performance of approaches.

\begin{table}[]
 \caption{Performances of different methods on the dataset.}
\begin{tabular}{ cccccc}
 \hline
\textbf{Method}            & \textbf{Precision} & \textbf{Recall} & \textbf{F1-score} & \textbf{AUC}   \\
 \hline
\textbf{Random Forest}    & 0.867     & 0.403  & 0.550    & 0.685 \\
\textbf{LightGBM}         & 0.892     & 0.416  & 0.567    & 0.686 \\
 \hline
\textbf{GraphSAGE}        & 0.973     & 0.545  & 0.699    & 0.785 \\
 \hline
\textbf{BiLSTM} & 0.966     & 0.480  & 0.641    & 0.755 \\
\textbf{TextCNN}          & 0.981     & 0.604  & 0.747    & 0.804 \\
\textbf{BERT}             & 0.984     & 0.619  & 0.760    & 0.861 \\
 \hline
\textbf{MCCF}             & \textbf{0.987}     & \textbf{0.854}  & \textbf{0.916}    & \textbf{0.933}\\
 \hline
\end{tabular}
 \label{table:different Method Table}
\end{table}
 \begin{figure*}[htb]
    \centering
    \includegraphics[width=14.7cm]{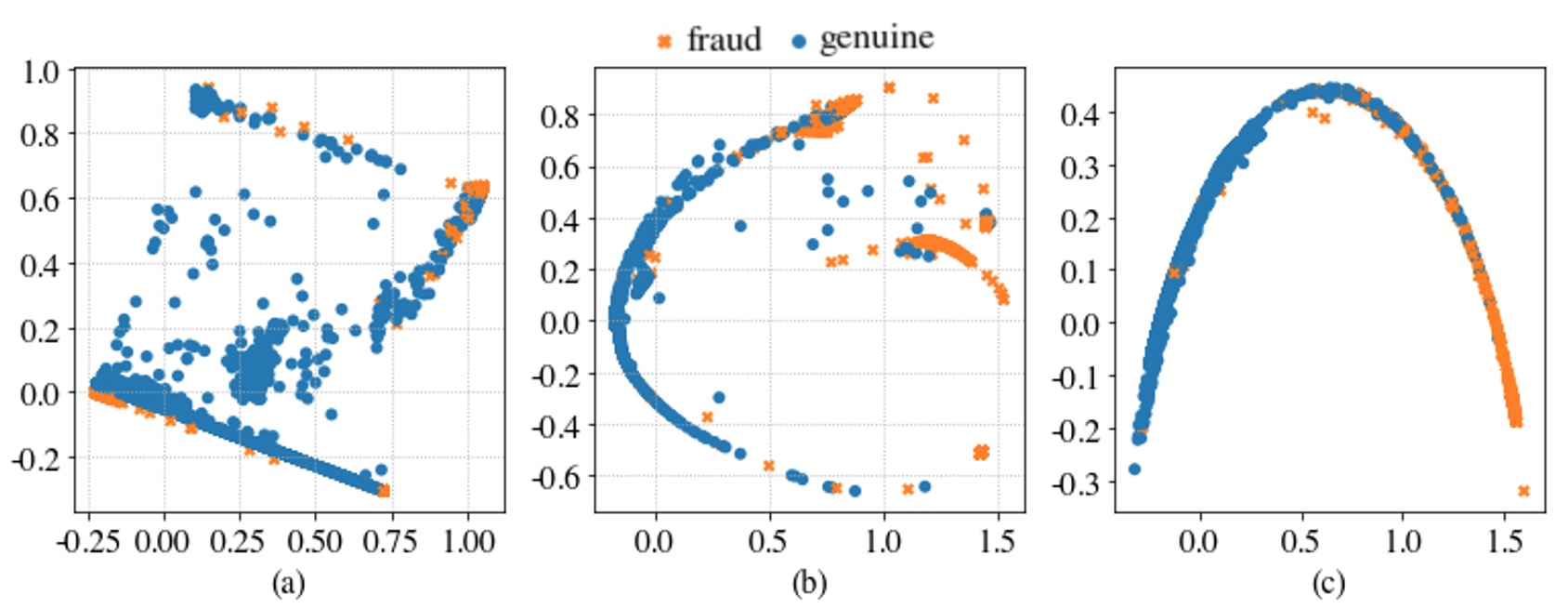}
    \caption{PCA projections: (a) original data; (b) last hidden layer with cross-entropy loss (MCCF$_{CE}$); (c) last hidden layer with NT-Xent loss (MCCF).}
    \label{fig:embedding visual}
\end{figure*}
\subsection{Main Results}
Table~\ref{table:different Method Table} demonstrates the main results of all compared methods on the dataset. The major findings from the experimental results can be summarized as follows:

(1) We can clearly observe that our model MCCF outperforms all the baselines by a large margin. Its F1-score, with reported value of 0.916, is at least 21.7\% higher than tree-based and graph-based methods, and AUC gets 14.8\% higher. Furthermore, MCCF is more advanced than sequence-based methods (i.e., BiLSTM, TextCNN and BERT), with at least 15.6\% increased F1-score and 7.2\% increased AUC. That is, the usage of sequence information and the further exploring on multimodal features make it more superior to the competitors. Besides, the obvious improvement of F1-score indicates that the model can detect more top-ranked click fraudsters under the same precision. This is critical to the real-world system when leveraging the business effect and interception rate.

(2) For baselines, LightGBM gets better performances than Random Forest among the tree-based methods. It achieves better F1-score due to deeper modeling residuals. BERT gets better performances than TextCNN and BiLSTM among the sequence-based methods. It achieves better F1-score via extracting different semantic information at different levels of abstraction. Moreover, it can be further seen that the graph-based method is more effective than tree-based methods via aggregating the statistical features of multiple media of the same user. F1-score is increased by more than 13.2\% and AUC is improved by 9.9\%. In addition, we observe that the sequence-based methods, e.g., TextCNN and BERT, are more effective than GraphSAGE due to taking advantage of behavioral information. F1-score is increased by more than 6.1\% and AUC is improved by 7.6\%.

\subsection{Ablation Test}
Furthermore, we perform the ablation test for our MCCF, and the results are shown in Table~\ref{table:ablation table}.
\subsubsection{The effects of modals.} Firstly, we demonstrate the effectiveness of different modals by removing the corresponding modal information (e.g., removing the behavior sequence) respectively. Compared the second to the fourth rows with the last row in Table~\ref{table:ablation table}, we can clearly see that all metrics get worse by removing any modal-specific information. It is the worst by removing behavior sequence, which means behavior sequence has more significant impact on detecting click fraud in our dataset. The results reflect the importance of macroscopically modeling multiple modals as well, since every modal has a positive contribution for our task.
\begin{table}[]
\caption{Performances of ablation test on the proposed MCCF method.}
\begin{tabular}{ lccccc}
 \hline
\textbf{Model}                  & \textbf{Precision} & \textbf{Recall} & \textbf{F1-score} & \textbf{AUC}   \\
 \hline
\textbf{MCCF}$_{\backslash B}$  & 0.970     & 0.735  & 0.836    & 0.856 \\
\textbf{MCCF}$_{\backslash V}$  & 0.975     & 0.776  & 0.864    & 0.882 \\
\textbf{MCCF}$_{\backslash WD}$ & 0.979     & 0.807  & 0.884    & 0.905 \\
\hline
\textbf{MCCF}$_{CE}$ & 0.985     & 0.832  & 0.902    & 0.918 \\
\hline
\textbf{MCCF}                   & \textbf{0.987}     & \textbf{0.854}  & \textbf{0.916}    & \textbf{0.933}\\
 \hline
\end{tabular}
 
 \label{table:ablation table}
\end{table}
\subsubsection{The effects of contrastive learning.} Next, to further verify the importance of contrastive learning in model integration and training, we take a comparison with our approach and its variants, as shown in Table~\ref{table:ablation table}. The variant {MCCF}$_{CE}$ is to change the loss function of the model from contrastive learning NT-Xent to cross entropy. We could clearly observe that {MCCF}$_{CE}$ performs worse than our full model, which illustrates the contrastive learning NT-Xent is more effective in optimizing imbalanced problem. Meanwhile, its decreasing values (v.s. MCCF) reflect that contrastive learning plays a significant role in click fraud detection. In addition, {MCCF}$_{CE}$ still gets a performance of 0.902 on F1-score and 0.918 on AUC and clearly outperforms BERT, which indicates the effectiveness of multimodal information in click fraud detection.

\subsection{Visualization}
We next look closer to the data, and visualize the principal components of original data and last hidden layer of model, as shown in Figure~\ref{fig:embedding visual}. The two axes represent two principal components analysis ~\cite{dunteman1989principal,smith2002tutorial} of data respectively.
It can be seen that the two principal components of the original data cannot distinguish between fraud and genuine clicks at all. Compared with the principal components of the original data, the system based on the cross entropy loss function, i.e., {MCCF}$_{CE}$, can clearly distinguish between fraud and genuine clicks. Furthermore, our MCCF model based on the contrastive learning loss function NT-Xent has better discrimination than {MCCF}$_{CE}$. From the visualizations, we demonstrate again that our MCCF, which incorporating multimodal information and contrastive learning, is effective in 
click fraud detection.

\section{CONCLUSIONS}
Advertising click fraud detection plays one of the vital roles in current E-commerce websites.
In this paper, we proposed the MCCF model that jointly exploits multimodal information network and contrastive learning for click fraud detection. We carefully analyzed the differences between fraudsters and genuine users in the advertising click scenario on statistical, behavioral and media relation information. The observations motivate the three essential modules in MCCF, extracting features from different perspectives separately. These 
three modules are integrated and jointly trained via contrastive learning. The experimental results on a real-world click fraud detection task show that our approach achieves promising performance and significantly outperforms the SOTA methods.
%In the future, we will explore the use of embedding of behavioral sequence and statistical features in the nodes of heterogeneous graph neural networks.

\begin{acks}
We would like to thank all the anonymous reviewers for their thoughtful and constructive comments and suggestions.
\end{acks}

%\printbibliography %Prints bibliography

\bibliographystyle{ACM-Reference-Format}
\bibliography{MCCF}

\end{document}